\PassOptionsToPackage{usenames,dvipsnames}{xcolor}
\documentclass{bmvc2k}
\usepackage{algorithm2e}
\usepackage{amsfonts}
\usepackage[export]{adjustbox}
\newsavebox\mybox
\usepackage{graphicx}
\usepackage{multirow}
\usepackage{longtable}
\usepackage{float}
\usepackage{subcaption}
\usepackage{adjustbox}
\usepackage[font={small}]{caption}
\usepackage{xcolor}

\title{PhysFlow: Skin tone transfer for remote heart rate estimation through conditional normalizing flows}

\addauthor{Joaquim Comas }{joaquim.comas@upf.edu}{1}
\addauthor{Antonia Alomar}{antonia.alomar@upf.edu}{1}
\addauthor{Adrià Ruiz}{adriaruiz@seedtag.com}{2}
\addauthor{Federico Sukno}{federico.sukno@upf.edu}{1}

\addinstitution{
Department of Information and Communication Technologies  \\
Pompeu Fabra University \\
Barcelona, Spain
}
\addinstitution{
 Seedtag\\
 Madrid, Spain
}

\runninghead{ COMAS ET AL.}{PhysFlow: Skin tone transfer for remote HR estimation}

\begin{document}

\maketitle

%%%%%%%%% ABSTRACT
\begin{abstract}

In recent years, deep learning methods have shown impressive results for camera-based remote physiological signal estimation, clearly surpassing traditional methods. However, the performance and generalization ability of Deep Neural Networks heavily depends on rich training data truly representing different factors of variation encountered in real applications. Unfortunately, many current remote photoplethysmography (rPPG) datasets lack diversity, particularly in darker skin tones, leading to biased performance of existing rPPG approaches. To mitigate this bias, we introduce PhysFlow, a novel method for augmenting skin tone diversity in remote heart rate estimation using conditional normalizing flows. PhysFlow adopts end-to-end training optimization, enabling simultaneous training of supervised rPPG approaches on both original and generated data. Additionally, we condition our model using CIELAB color space skin features directly extracted from the facial videos without the need for skin-tone labels. We validate PhysFlow on publicly available datasets, UCLA-rPPG and MMPD, demonstrating reduced heart rate error, particularly in dark skin tones. Furthermore, we demonstrate its versatility and adaptability across different data-driven rPPG methods.
\end{abstract}

%%%%%%%%% BODY TEXT
\section{Introduction}
\label{sec:intro}

%Since Takano et al. \cite{takano2007heart} and Verkruysse et al. \cite{verkruysse2008remote} demonstrated the feasibility of remote HR measurement from facial videos, researchers have proposed diverse methods for physiological data recovery. 

%Recently, the research community has increasingly focused on the camera-based measurement of human physiological signals and their potential applications  \cite{ronca2021video, benezeth2018remote, liu20163d}, particularly in the extraction and analysis of vital signs such as the heart rate (HR), heart rate variability (HRV), respiration rate (RR), oxygen saturation (SpO2), and blood volume pulse (BVP). Among these vital signs, HR has been the most extensively studied due to its relevance to health and human-computer interaction applications, e.g. by using HR to complement facial expressions for the analysis of human emotions \cite{jung2019utilizing,comas2020end}.

In recent years, the interest on camera-based measurement of physiological signals has undergone substantial growth caused by its potential applicability in clinical \cite{huang2021neonatal} and human-computer interaction applications \cite{benezeth2018remote, nowara2020near}. The recent advancements in deep learning have further propelled this field. However, data-driven methods often require a large amount of training data to achieve good generalization, making their performance %highly dependent on training and evaluation splits that may not reflect real-world scenarios. 
highly dependent on the distribution and scale of the dataset chosen for training. Additionally, many rPPG datasets suffer from biases, particularly regarding demographic diversity, leading to unfair performance depending on skin tone, especially for underrepresented ethnicities. 

Previous studies \cite{fallow2013influence, wang2015novel, nowara2020meta} have investigated the effects of demographic biases in existing rPPG datasets, revealing a significant decline in remote heart rate estimation performance for subjects with the darkest skin tones. While these studies highlighted the consequences of skin tone bias, none of them proposed concrete solutions to address the problem. Although the creation of datasets that accurately represent all skin types may seem like a simple solution to overcome this problem, it is often impractical and resource-intensive, especially during the recruiting of the participants. Consequently, each dataset typically over-represents the population of the country in which it was collected. For example, datasets like VIPL-HR \cite{niu2018vipl} or PURE \cite{stricker2014non} predominantly consist of Asian or Caucasian populations, respectively.

Only a few recent studies have addressed skin tone bias. Two of them \cite{mcduff2022scamps, wang2022synthetic} used synthetic avatars to tackle data imbalance. However, these approaches are often not photo-realistic and can be time-consuming in data generation. Another work \cite{vilesov2022blending} employed radar hardware to improve rPPG signal recovery for dark skin tones. Nevertheless, this method is not always feasible and limits rPPG versatility with conventional cameras. Finally, Ba et al. \cite{ba2022style} proposed a two-stage optimization using external pre-training \cite{yucer2020exploring} to transfer skin tone content, which is an interesting possibility to balance skin-tone distributions as long as external data with the underrepresented skin tones are available for pre-training. %. However, \cite{ba2022style} is constrained by limited external skin tone pre-training data. %and its two-stage optimization may lead to artifacts in skin tone transfer, resulting in certain facial regions being less modified than others.

%Inspired by the success of flow-based generative models in exact latent variable inference \cite{dinh2014nice, dinh2017density}, 
In this context, we propose a novel data augmentation approach leveraging normalizing flows \cite{dinh2014nice, dinh2017density} to disentangle the skin tone information from the rest of the facial video content. Additionally, 
they allow for attribute control by concatenating parameters to embeddings, facilitating the conditioning of specific properties such as skin tone. Furthermore, in this paper, we approach the representation of skin tone for rPPG estimation differently than before. Previous works typically used the Fitzpatrick scale \cite{fitzpatrick1988validity} to evaluate or annotate skin tone, categorizing it into six levels from I (lightest) to VI (darkest). In contrast, we propose using a bi-dimensional representation \cite{thong2023beyond} to quantify the apparent skin color, which extracts the perceptual light and hue angle from facial videos, offering three key advantages over the Fitzpatrick scale. Firstly, it enables automated skin tone measurement in each facial video, eliminating the need for manual annotations and allowing to use it into unlabeled data, common in the majority of rPPG datasets. Secondly, it simplifies the collection and annotation for new rPPG datasets, reducing human error inherent in manual Fitzpatrick scale annotations due to subjective perceptions. Lastly, unlike the Fitzpatrick scale, which considers only lightness, this bi-dimensional representation accounts for variations in hue, distinguishing between red and yellow hues. This broader perspective better accommodates diverse skin tones, particularly given that common experiments in existing rPPG datasets involve varying lighting conditions using different types of light sources.

The main contributions of the paper are three-fold: 
\begin{itemize}
    \item We introduce PhysFlow, a novel skin tone data augmentation approach for rPPG estimation, which adopts conditional normalizing flows to disentangle skin tone information from other appearance and temporal facial video features.
    \item We propose a novel training that leverages skin tone CIELAB color space representation without requiring external labels.
    \item We train the PhysFlow model end-to-end to optimize any supervised rPPG approach using the original and generated data within the same data generation process promoting a fast adaptation.
    
\end{itemize}
%1)  2)  3) We train the PhysFlow model end-to-end to optimize any supervised rPPG approach using the original and generated data within the same data generation process promoting a fast adaptation. \ar{I would use an itemize to have it as different bullet points. More readable when the reviewer have finished reading the paper and has to write the review.}
%\vspace{-1em}
\section{Related work}
\label{sec:related}
\textbf{Camera-based PPG measurement}. Since Takano et al. \cite{takano2007heart} and Verkruysse et al. \cite{verkruysse2008remote} demonstrated remote HR measurement feasibility from facial videos, diverse methods for physiological data recovery have emerged. Some focus on regions of interest, employing techniques like Blind Source Separation \cite{poh2010non, poh2010advancements, lewandowska2011measuring}, Normalized Least Mean Squares \cite{li2014remote}, or self-adaptive matrix completion \cite{tulyakov2016self}. Others utilize the skin optical reflection model, projecting RGB skin pixel channels into an optimized subspace to mitigate motion artifacts \cite{de2013robust, wang2015novel}. Deep learning-based methods \cite{vspetlik2018visual, yu2019remote, perepelkina2020hearttrack, lee2020meta, lu2021dual, nowara2021benefit} have surpassed conventional techniques, achieving state-of-the-art performance in estimating vital signs from facial videos. Some combine traditional methods with Convolutional Neural Networks (CNNs) to leverage advanced features \cite{niu2018synrhythm, niu2019rhythmnet, song2021pulsegan}. Other recent works \cite{lee2020meta, liu2021metaphys} explore unsupervised approaches using meta-learning, enhancing generalization in out-of-distribution cases. Unlike previous approaches, some propose end-to-end models \cite{chen2018deepphys, Yu2019RemotePS, perepelkina2020hearttrack} to directly extract the rPPG signal from facial videos. Transformer-based models like Physformer \cite{yu2023physformer++} and RADIANT \cite{gupta2023radiant} have gained attention for leveraging long-range spatiotemporal features, although they currently lack optimization for mobile deployment. While promising, they may not yet demonstrate a significant performance advantage over CNN-based models \cite{liu2023efficientphys}. Finally, works like \cite{liu2023efficientphys} and \cite{comas2022efficient} propose lightweight rPPG frameworks with competitive HR results while controlling computational cost.

\noindent \textbf{Skin tone bias in rPPG measurement}. Earlier works \cite{fallow2013influence, wang2015novel, addison2018video} noted the lowest signal-to-noise ratio and highest error rates for darker skin tones in assessments on private datasets. Prior to the work of Nowara et al. \cite{nowara2020meta}, the influence of skin tone on rPPG extraction had only been examined in limited datasets without population diversity and data-driven approaches. In their meta-analysis, Nowara et al. investigated the impact of skin type on rPPG signal recovery for each Fitzpatrick skin tone using three datasets. They found significantly poorer rPPG signal recovery for subjects with category VI compared to other skin types. In \cite{ernst2021optimal}, a new color space for rPPG estimation was proposed, showing consistent results across all skin tones, albeit with a performance drop for darker skin tones (V and VI), similar to \cite{nowara2020meta}. Subsequently, Dasari et al. \cite{dasari2021evaluation} conducted an extensive study analyzing estimation biases of camera PPG methods across diverse demographics. They observed similar biases as with contact-based devices and environmental conditions. %Additionally, they evaluated the impact of different color spaces, finding that those capable of differentiating luminance from chrominance information tend to effectively preserve chrominance data, making them suitable for rPPG estimation even in scenarios with fluctuating background lighting and brightness.

\noindent \textbf{rPPG data-augmented approaches}. While various works have identified skin tone bias in rPPG estimation, only a few solutions have directly addressed its mitigation. Recent studies have proposed different data augmentation approaches considering factors like motion \cite{paruchuri2024motion} or data scarcity \cite{tsou2020multi, hsieh2022augmentation}, but without targeting skin tone bias. Two recent works aimed to balance demographic groups present in current rPPG datasets. Firstly, Vilesov et al. \cite{vilesov2022blending} proposed a method to mitigate bias across skin tones using multi-modal fusion, combining information from an RGB camera and 77 GHz radar. However, the need for a radar device limits the applicability of camera-based measurements in all scenarios. Alternatively, Ba et al. \cite{ba2022style} introduced a two-step joint optimization framework to translate light-skinned to dark-skinned facial videos while maintaining their pulsatile signals, resulting in a considerable reduction of HR estimation error in their private dataset. However, this approach has two main limitations. Firstly, the use of a predefined model \cite{yucer2020exploring} to translate skin tone is limited to only four labels (African, Asian, Caucasian, and Indian) and may not discern different variations for each skin label. Secondly, the two-stage joint optimization can lead to unrealistic skin tone transformations, with certain facial regions appearing more illuminated than others. In contrast, our novel approach is independent of external pre-training and can translate skin tone from the same data or a reduced set of data. Our end-to-end approach using conditional normalizing flows produces more homogeneous, realistic, and diverse skin tone transfers. Also some studies have tackled the limited availability of data in existing benchmarks by generating synthetic data. For example, McDuff et al. \cite{mcduff2022scamps} used Blender to create synthetic avatars aligned with cardiac and respiratory signals across various scenarios, while Wang et al. \cite{wang2022synthetic} employed a 3D morphable face model to generate synthetic facial videos. Although both methods show promise, they have limitations. Primarily, the generation of synthetic data can be excessively time-consuming, especially in the case of \cite{mcduff2022scamps}, and they may suffer from domain shift, where models trained on synthetic data fail to perform well on real datasets due to the domain gap between the datasets, requiring a domain adaptation stage \cite{wang2018deep}. Additionally, in both instances, the facial appearance lacks photo-realism, and certain physiological factors like Pulse Transit Time (PTT) are not considered. In contrast, our data augmentation approach proposes direct training that simultaneously trains the rPPG model while generating new data, preserving the photo-realism of the original data and the characteristics of pulsatile data.

\section{Methodology}
\label{sec:method}
%In this section, we introduce and define the conditioned normalizing flows. Subsequently, we present our proposed end-to-end skin tone transfer approach and finally present our optimized objective function for remote HR estimation.
\subsection{Preliminaries: Continous Normalizing Flows}
%\textbf{Discrete Normalizing Flows.} 
Normalizing flows (NFs) \cite{rezende2015variational, tabak2010density} are a class of generative neural networks that estimate an unknown complex data distribution, $p(z)$, from a known and simpler base probability distribution, $p(u)$, such as a normal distribution $\mathcal{N}(\mu,\sigma^{2})$. The mapping between both distributions is achieved by applying a sequence of bijective and differentiable transformations $f_1, ...,f_K : \mathbb{R}^{d} \rightarrow \mathbb{R}^{d} $. Additionally, the mapping between both distributions can be constrained to some set of conditions or attributes $c$, these types of NFs are commonly known as conditional normalizing flows (c-NFs). Hence, the relationship between $z|c$ and $u|c$ can be defined as:
\begin{equation}
    u= f_{\theta}(z;c),   z= f^{-1}_{\theta}(u;c)
\end{equation}

\noindent where $u\sim p(u|c)$, $z\sim p(z|c)$, and $f_\theta$ is a bijective neural network parameterized by $\theta$.  Mathematically, c-NFs can express the log-likelihood by applying the change of variables rule:  
\begin{equation}
    \log p(z; c, \theta) = \log p(f_{\theta}(z;c)) - \sum_{i=1}^{k-1} \log | det(J_{\theta_i}(z; c)|
\end{equation}
\noindent where  $J_{\theta_i}(z; c)=\frac{\partial f_{\theta_i}(z;c)}{\partial z}$ is the Jacobian, and $|det(J_{\theta_i}(z; c)|$ represents the volume change caused by the transformations of $f_\theta$. 

The normalizing flow can be generalized from a discrete sequence to a continuous transformation \cite{chen2018neural, grathwohl2018scalable}, parameterizing the dynamics of data transformation over time $\frac{\partial u_\tau}{\partial \tau} = g_{\theta}(\tau,c,u_\tau)$, where $g_{\theta}$ is a neural network, $u_\tau$ is the state at time $\tau$, $u \sim \mathcal{N}(\mu,\sigma^{2})$ and $c$ is the conditioning attributes. The Continuous Normalizing Flow (CNF) computes $z$ from $u$ by integrating $g_\theta$ over time, express as:

\begin{equation}
   z = u_{\tau_0} + \int_{\tau_0}^{\tau_1} g_{\theta}(\tau,c,u_\tau) d\tau
\end{equation}

Following the Instantaneous Change of Variables theorem \cite{chen2018neural} where the change in log probability of a continuous random variable is equal to the trace of the Jacobin matrix, the total change in log density can be written as: 
\begin{equation}
    \log p(u(\tau_1)) = \log p(u(\tau_0)) - \int_{\tau_0}^{\tau_1} Tr\left(\frac{\partial g_{\theta}}{\partial u_\tau}\right) d\tau
    \label{log_nf}
\end{equation}

To compute the gradients with respect to the parameters a black-box ordinary differential equation (ODE) solver is applied using an adjoint sensitivity method \cite{pontryagin2018mathematical}. In this work, we adopt CNFs because they are demonstrated to have better expressiveness and versatility than discrete NFs \cite{grathwohl2018scalable, abdal2021styleflow}. 
%Flow-based models offer several advantages, including ease of sampling, stable training, and accurate probability density estimation compared to other generative models like GANs \cite{goodfellow2014generative} or VAEs \cite{kingma2013auto}.

\subsection{PhysFlow}
\label{subsec:physflow}
\begin{figure}[h]
\centering
\includegraphics[width=0.95\textwidth,height=0.50\textwidth]{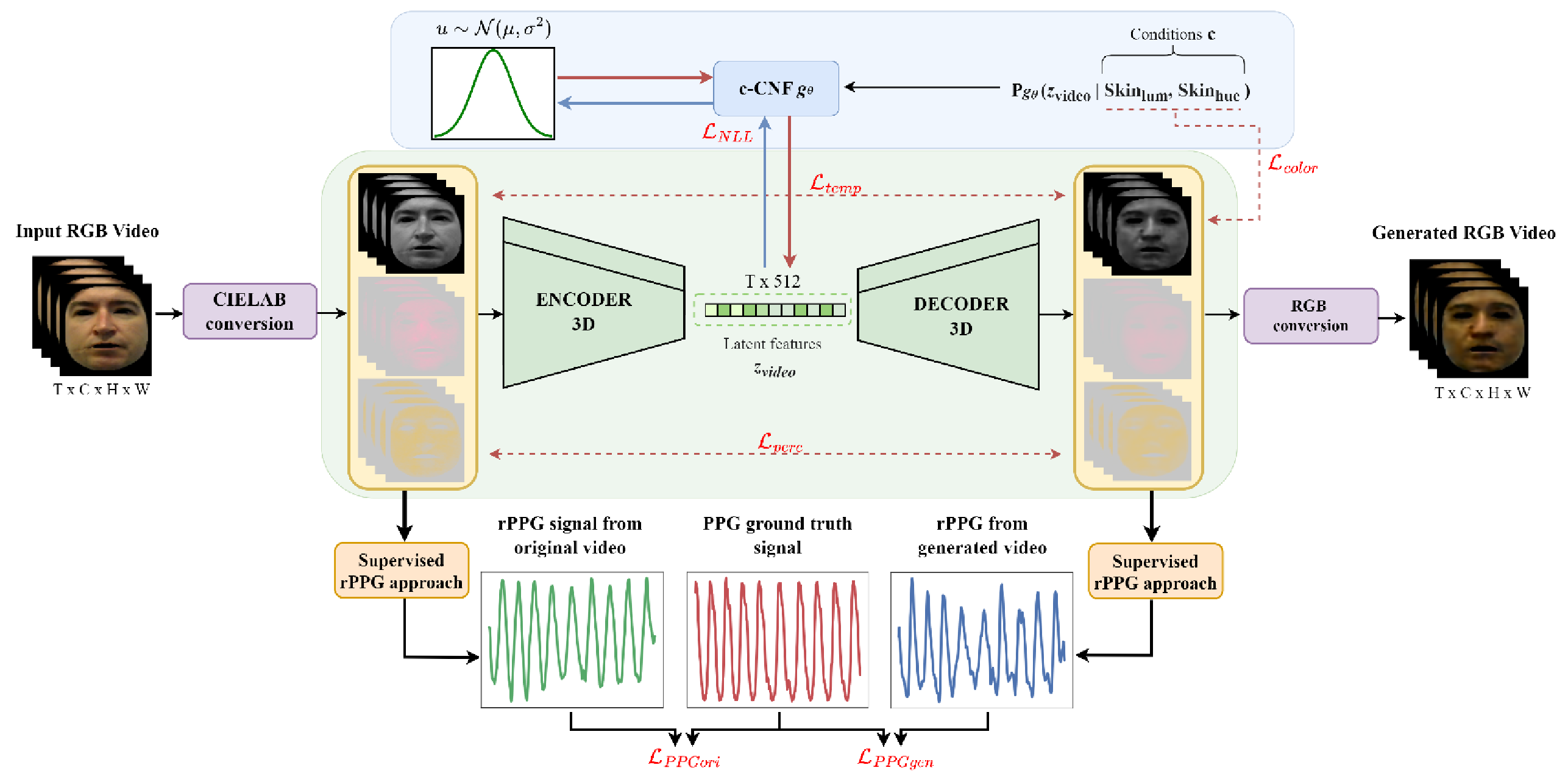}

    \caption{PhysFlow pipeline: A 3D-CNN AE encodes entangled video facial content into a latent embedding. This embedding is then processed by c-CNFs to disentangle the skin tone content. Simultaneously, the rPPG model is iteratively trained using both original and skin tone-augmented data.}
    \label{fig:Physflow}

\end{figure}

%\ar{I would write here 3 lines reminding what is the purpose of we want to do (e.g, data augmentation + rppg estimation) and outline the purpose of the different components that will be presented in the next subsections.}
The proposed PhysFlow approach, outlined in Fig.\ref{fig:Physflow}, aims to mitigate the effect of skin tone bias in rPPG data-driven models by generating a diverse range of skin tone-augmented facial videos through the usage of CIELAB skin tone features. Our model comprises three main components: a 3D auto-encoder (AE), a conditional CNF network, and the rPPG model.

\subsubsection{3D video reconstruction}
\label{sub:3dae}
Inspired by several flow-based generative models \cite{lippe2022citris, li2022style} in different computer vision tasks, we employ a conventional auto-encoder model, which is trained to encode and decode the high-dimensional information into low-dimensional feature embedding, independently of any disentanglement. Unlike previous works, we aim to preserve the underlying physiological signals that depend on subtle spatio-temporal variations. Therefore, the preservation of inter-frame content is crucial to recover the rPPG signal, thus we choose a 3D-CNN AE to model the spatio-temporal information of the input facial videos (the architecture details are explained in supplementary material). During the 3D-CNN AE training, the features are only spatially downsampled and upsampled in order to not degrade the temporal content. To train the proposed 3D-CNN AE, we utilize the following combined loss:  

\begin{equation}
    \mathcal{L}_\mathrm{{AE}} = \alpha_\mathrm{{1}} \mathcal{L}_\mathrm{{rec}} + \alpha_\mathrm{{2}} \mathcal{L}_\mathrm{{perc}} +\alpha_\mathrm{{3}}  \mathcal{L}_\mathrm{{rPPG}},
    \label{ae_loss}
\end{equation}
\qquad
    \begin{equation}
        \mathcal{L}_\mathrm{{rec}} = \big\| \hat{X}_t- X_t \big\|_{1}, 
        \mathcal{L}_\mathrm{{perc}} =  \frac{1}{C_j H_j W_j} \big\| \phi_j(\hat{X}_t))-\phi_j(X_t)) \big\|_{1}, 
        \mathcal{L}_\mathrm{{rPPG}} =  \big\| rPPG_{rec}-rPPG_{ori} \big\|_{2} 
    \end{equation}

\noindent where $\alpha_\mathrm{{*}}$, controls the importance between different loss
components and $\mathcal{L}_{rec}$ is the reconstructed loss between the reconstructed $\hat{X}_t$ and original $X_t$ facial videos computed by the $L_{1}$-loss. The $\mathcal{L}_{perc}$ is a VGG perceptual loss \cite{johnson2016perceptual} adopted to retain high-frequency details between the reconstructed and original video, where $\phi_j$ represents the perceptual function which outputs the activation function of the $j$th layer in the VGG network and $C_j$, $H_j$, $W_j$ are the dimensions of the tensor feature map. Finally, $\mathcal{L}_{rPPG}$ is a $L_{2}$-loss between the extracted rPPG in the original and reconstructed facial video to ensure the preservation of the physiological data.

\subsubsection{Skin tone representation and conditioning}
\label{sub:st_cond}
Our data augmentation approach relies on two primary assumptions. Firstly, for effective augmentation, the unbalanced dataset should feature a minimum level of skin tone diversity, enabling the transfer of skin tone without relying on an external dataset. Secondly, consistent luminance conditions within the same video are necessary for successful skin tone transfer. Hence, in this work, we select the UCLA-rPPG dataset \cite{wang2022synthetic}. Despite its significant imbalance in skin tone representation, the dataset offers sufficient diversity and samples, while maintaining constant illumination across each facial video. %\adria{Have not you mention that robustness to illumination changes is an advantage of your method?}  
Based on these assumptions, we model our skin tone transfer employing a bi-dimensional representation \cite{thong2023beyond} expressed in CIELAB color space through two components: luminance and hue. The luminance component determines the lightness or darkness of the skin, while the hue component, derived by dividing the alpha and beta channels, spans from red to yellow. Fig. \ref{fig:ucla_labels} illustrates the comparison between our bi-dimensional skin tone label representation and the annotated Fitzpatrick labels from the UCLA-rPPG dataset. We notice that the luminance and hue values are not well captured by the subjective Fitzpatrick annotations, i.e. some subjects annotated in a specific skin type may belong to another one, particularly between skin types I to V, as seen in Fig. \ref{fig:ucla_labels}. %similarities in luminance and hue attributes among subjects across different Fitzpatrick scales, particularly skin types I-V. Additionally, the visual examples underscore human error annotation, as some subjects annotated in a specific skin type may belong to others. 
These subjective annotations may stem from the challenge of labelling a diverse range of skin tones within only six main classes, where considerable diversity exists within each class. Therefore, the use of CIELAB skin tone labels not only facilitates the skin tone transfer without relying on manual Fitzpatrick labels but also ensures the ability to map large variations of skin tones and overcome illumination changes between different trials on the same subject, which could affect the perception of skin type.

To apply skin tone transfer, once the 3D auto-encoder has converged, we freeze its parameters and proceed to train the whole system. During the end-to-end training, the c-CNF module is responsible for mapping the entangled latent representation to a disentangled form. The invertibility of the flow-based model ensures that no information is lost during the transition from the entangled to the disentangled latent space. To achieve skin tone disentanglement, our CNF model is conditioned on luminance and hue values computed for a sequence of frames. %Additionally, inspired by \cite{abdal2021styleflow}, we decided to include two extra attributes: glasses and beard. By incorporating these attributes, the joint attribute training aims to learn a stable conditional embedding without associating any of these features exclusively with a particular skin tone. This approach helps mitigate potential biases caused by data imbalance, such as the glasses or beard attribute could be erroneously linked to a specific skin tone due to its underrepresentation.

\savebox{\mybox}{\includegraphics[width=5.5cm,height=4.3cm]{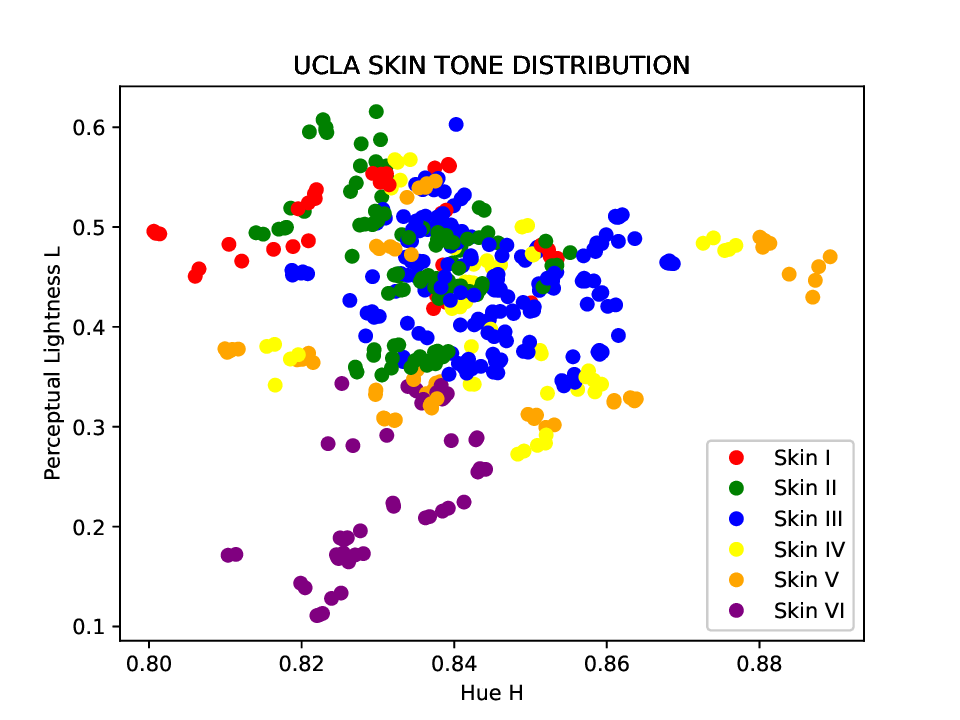}}
\begin{figure}[t]
    \centering
    \begin{minipage}{0.4\textwidth}
        \centering
        \usebox{\mybox}
    \end{minipage}
    \begin{minipage}{0.4\textwidth}
        \centering
        \vbox to \ht\mybox{%
            \vfill
\includegraphics[width=5cm,height=3.5cm]{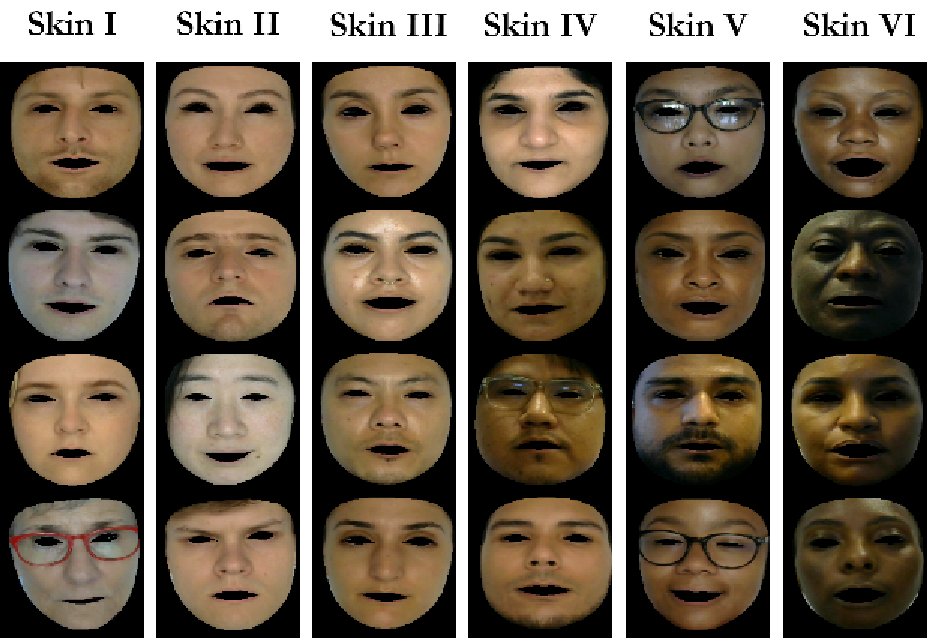}
            \vfill
        }   
    \end{minipage}
\caption{Skin tone representation in UCLA-rPPG. Left: Distribution representation of skin tone in terms of CIELAB luminance and hue compared to annotated Fitzpatrick scale labels from the dataset. Right: Visual examples of different skin tones with Fitzpatrick labels.}
\label{fig:ucla_labels}

\end{figure}

\subsubsection{Model training and optimization}

For training the PhysFlow model, we utilize end-to-end training, optimizing both the c-CNF module and the rPPG approach simultaneously. This involves minimizing the continuous form of the negative log-likelihood, as previously introduced in Equation \ref{log_nf}:

\begin{equation}
\mathcal{L}_\mathrm{{NLL}} =  - \log p(u(\tau_0)) + \int_{\tau_0}^{\tau_1} Tr\left(\frac{\partial g_{\theta}}{\partial u_\tau}\right) d\tau
\end{equation}

\noindent During the PhysFlow training, we also control the synthesis of the skin tone transfer by conditioning the c-CNF with specific transferable skin tone values and assessing the generated output. For that, we introduce three regularization losses. Firstly, we compute a consistency color loss $\mathcal{L}_\mathrm{{color}}$ between the conditioned skin tone values $st_{cond}$ and the resulting skin tone measured from the generated video $\hat{st}_{pred}$. Secondly, to enforce the preservation of the temporal content we consider a temporal consistency loss $\mathcal{L}_\mathrm{{temp}}$, which is computed by minimizing the error of the first-order derivative of the original $z_{video}$ and the conditioned $\hat{z}_{video}$ latent space:   

\begin{equation}
    \mathcal{L}_\mathrm{{color}} = \big\|\hat{st}_{pred} - st_{cond} \big\|_{1}, 
    \qquad
    \mathcal{L}_\mathrm{{temp}} =  \Big\| \frac{\partial \hat{z}_{video}}{\partial t} - \frac{\partial z_{video}}{\partial t} \Big\|_{1}
\end{equation}

\noindent Lastly, we use the same perceptual loss $\mathcal{L}_\mathrm{{perc}}$ from Eq. \ref{ae_loss} to keep the overall appearance of the original face. %force similar facial appearance as the original face. 
At the same time, to optimize the supervised rPPG approach we incorporate a physiological loss which combines the minimization of the rPPG signal concerning the ground truth PPG from the original and also from the generated video, which can be written as:

\begin{equation}
\mathcal{L}_\mathrm{{phys}} = \mathcal{L}_\mathrm{{PPGgen}} +\mathcal{L}_\mathrm{{PPGori}}   
\end{equation}
To compute $\mathcal{L}_\mathrm{{phys}}$, in our experiments, we use the Negative Pearson correlation \cite{Yu2019RemotePS} and TALOS \cite{comas2022efficient} losses depending on the selected rPPG approach. Finally, we can express the overall training objective for PhysFlow as the minimization of the sum of all the losses: 

\begin{equation}
    \mathcal{L}_\mathrm{{overall}} = \lambda_\mathrm{{1}} \mathcal{L}_\mathrm{{NLL}} + \lambda_\mathrm{{2}} \mathcal{L}_\mathrm{{perc}} +\lambda_\mathrm{{3}} \mathcal{L}_\mathrm{{color}} +\lambda_\mathrm{{4}} \mathcal{L}_\mathrm{{temp}} +\lambda_\mathrm{{5}} \mathcal{L}_\mathrm{{phys}}  
\end{equation}

\noindent where $\lambda_\mathrm{{*}}$, controls the importance between different loss components. %For fast convergence, we first pre-trained the c-CNF module and then trained end-to-end jointly with the supervised rPPG model, while 3D auto-encoder remains frozen as explained in the subsubsection. \ref{sub:st_cond}. 

%An end-to-end training is performed optimizing the c-CNF module and the rPPG approach while the 3D auto-encoder remains frozen.

\section{Experiments}
\label{sec:experiments}

\subsection{Experimental Setup}
\textbf{Datasets and models:} We consider two recent datasets, UCLA-rPPG \cite{wang2022synthetic} and MMPD \cite{tang2023mmpd}. Both benchmarks consist of facial videos and corresponding gold-standard PPG signals, while also containing Fitzpatrick labels  (see supplementary materials for more details). To evaluate PhysFlow performance, we use three deep-learning models: PhysNet \cite{Yu2019RemotePS}, EfficientPhys \cite{liu2023efficientphys}, and DPMN+TDM \cite{comas2024pulse}. For our experiments, we use the UCLA-rPPG dataset for training and MMPD for cross-dataset evaluation. Also, we include some traditional methods such as POS \cite{wang2016algorithmic} and CHROM \cite{de2013robust} as baseline results.

%\textbf{Datasets:} The UCLA-rPPG dataset \cite{wang2022synthetic} comprises 489 videos from 98 subjects with diverse characteristics, including skin tones, ages, genders, and ethnicities. It was specifically curated to generate synthetic avatars for rPPG estimation. Each subject underwent five trials, with each trial lasting approximately 1 minute. The recordings were captured at a resolution of $640\times480$ pixels and 30 frames per second (FPS), in an uncompressed format. Synchronous gold-standard PPG and HR measurements were collected alongside the facial videos. The Multi-domain Mobile Video Physiology Dataset (MMPD) \cite{tang2023mmpd} comprises 660 one-minute videos from 33 subjects, totaling 11 hours of recordings from mobile phones. Each subject contains twenty trials to capture variations in skin tone, body motion, and lighting conditions. These trials included stationary, rotation, talking, and walking tasks performed under four different lighting conditions. The videos were recorded at 30 FPS with a resolution of 1280x720 pixels, while PPG signals were simultaneously recorded using an HKG-07C+ oximeter at 200Hz and downsampled to 30Hz. Additionally, each subject was provided with eight descriptive labels, including the Fitzpatrick scale.

\textbf{Metrics and evaluation protocol:} To evaluate the HR estimation performance of the proposed model, we adopt the same metrics used in the literature, such as the mean absolute HR error (MAE), the root mean squared HR error (RMSE), the mean absolute percentage error (MAPE) and Pearson’s correlation coefficients R \cite{li2014remote, liu2021metaphys, liu2023efficientphys}. Our experiments are performed using subject evaluation with a 10-sec sequences average with no overlap to compute HR estimation for all reported metrics. For our cross-dataset experiment, we utilize the MMPD dataset considering only the skin tone. For this reason, we followed the same protocol in \cite{tang2023mmpd} to exclude from our analysis other factors such as exercise or lighting conditions. 

\textbf{Implementation details: } In all our experiments, we utilize the Mediapipe Face Mesh model \cite{lugaresi2019mediapipe, kartynnik2019real} to focus our analysis only on facial skin pixels. After masking the facial video, each frame is resized to $96\times 96$ pixels. The PPG ground truth is pre-processed following \cite{dall2020prediction} to denoise the raw PPG signal, which facilitates a better model convergence during the training. We use PyTorch 2.1.2 \cite{paszke2019pytorch} and train on a single NVIDIA RTX3060 using sequences of 300 frames without overlap and AdamW optimizer with a learning rate of 0.0001. As explained in section \ref{sub:st_cond},  we first pre-trained the 3D AE for 310 epochs and then froze the weights. Following \cite{lippe2022citris}, during the 3D AE training, we add a small Gaussian noise of 0.05 to the latent space to prevent the latent distribution from collapsing to single delta peaks. Furthermore, for fast convergence, once the 3D AE is trained we also pre-trained the c-CNF module for 50 epochs and then trained end-to-end jointly with the supervised rPPG model. The weights for different losses in 3D AE training are set as $\alpha=500$, $\alpha=10$ and $\alpha=5$, while in PhysFlow training are set as $\lambda_1=1$, $\lambda_2=100$, $\lambda_3=1000$, $\lambda_4=0.0001$ and $\lambda_5=10$.
Finally, the predicted rPPG signal is filtered using a band-pass filter with cut-offs of 0.66-3Hz while the heart rate is calculated using Chirp-Z Transform (CZT) \cite{comas2024deep}.

\subsection{Experimental Results}
%\vspace{-0.5em}
Table \ref{table:baseline} summarizes the assessment of remote HR performance across different Fitzpatrick scale splits in MMPD. These baseline evaluations are conducted using both traditional and data-driven approaches. From these results, several conclusions can be drawn. Firstly, across all methods, there is a noticeable decline in heart rate performance with darker skin tones, consistent with findings in previous studies \cite{addison2018video, ernst2021optimal, nowara2020meta}. Despite the similar representation of skin types IV, V, and VI between the datasets, a significant performance gap is evident, potentially influenced by their proximity to the more prevalent types such as skin tones II and III (as depicted in supplementary materials). Finally, deep learning models demonstrate substantial superiority over traditional approaches for lighter skin types (III and IV). However, for the most challenging skin types (V and VI), certain traditional methods perform comparably or even outperform some deep learning models, indicating skin tone-biased performance attributed to dataset imbalance.

\begin{table}[t]
  %\captionsetup{font=footnotesize}
  \renewcommand{\arraystretch}{1.5}
  \centering
  \adjustbox{width=\textwidth}{
  \large
  \begin{tabular}{c|ccc|ccc|ccc|ccc|ccc}
    \hline
    \multicolumn{1}{c}{ } & \multicolumn{6}{|c}{Hand-crafted methods} &\multicolumn{9}{|c}{Deep learning methods}
    \\
    \hline
    \multirow{1}{*}{Method} & \multicolumn{3}{c|}{POS \cite{wang2016algorithmic}} & \multicolumn{3}{c|}{CHROM \cite{de2013robust}} & \multicolumn{3}{c|}{PhysNet \cite{Yu2019RemotePS}} & \multicolumn{3}{c|}{EfficientPhys \cite{liu2023efficientphys}} & \multicolumn{3}{c}{DPMN+TDM \cite{comas2024pulse}} 
    \\
    \hline
    Skin Tone &
     MAE$\downarrow$ & RMSE$\downarrow$ & R$\uparrow$& MAE$\downarrow$ & RMSE$\downarrow$  & R$\uparrow$& MAE$\downarrow$ & RMSE$\downarrow$  & R$\uparrow$& MAE$\downarrow$ & RMSE$\downarrow$ & R$\uparrow$& MAE$\downarrow$ & RMSE$\downarrow$ & R$\uparrow$\\ \hline
    \multicolumn{16}{c}{\textbf{Lighter skin types}}
    \\
    \hline
    III & 5.76 & 9.67 & 0.48 & 6.57 & 10.46 & 0.33  & 3.90 & 7.98  & 0.56 & 4.18 & 9.40 & 0.60 & 3.09 & 5.92 & 0.78  \\
    IV & 9.06 & 13.51 & 0.23 &  10.57 & 13.80 & 0.15 & 7.69 & 10.93 & 0.56  & 7.42 & 14.35 & 0.41 & 4.98 & 8.78 & 0.69   \\
    \hline
    \multicolumn{16}{c}{\textbf{Darker skin types}}
    \\
    \hline
    V & 12.78 & 16.69 & -0.03 & 14.65 & 18.91 & -0.12 & 10.65 & 13.73 & 0.16 & 11.92 & 19.84 & 0.04 & 7.77 & 11.90 & 0.46  \\
    VI & 11.17 & 15.34 & 0.26 & 12.53 & 16.47 & 0.06 & 13.74 & 18.19 & -0.04 & 16.25 & 23.84 & 0.05 & 11.08 & 15.22 & 0.25  \\

    \hline
  \end{tabular}
  }
   \caption{Baseline cross-database evaluation on  MMPD dataset for each skin tone (beats per minute).}
  \label{table:baseline}
    \vspace{-1em}
\end{table}

\begin{figure}[b]

\centering    \includegraphics[width=\textwidth]{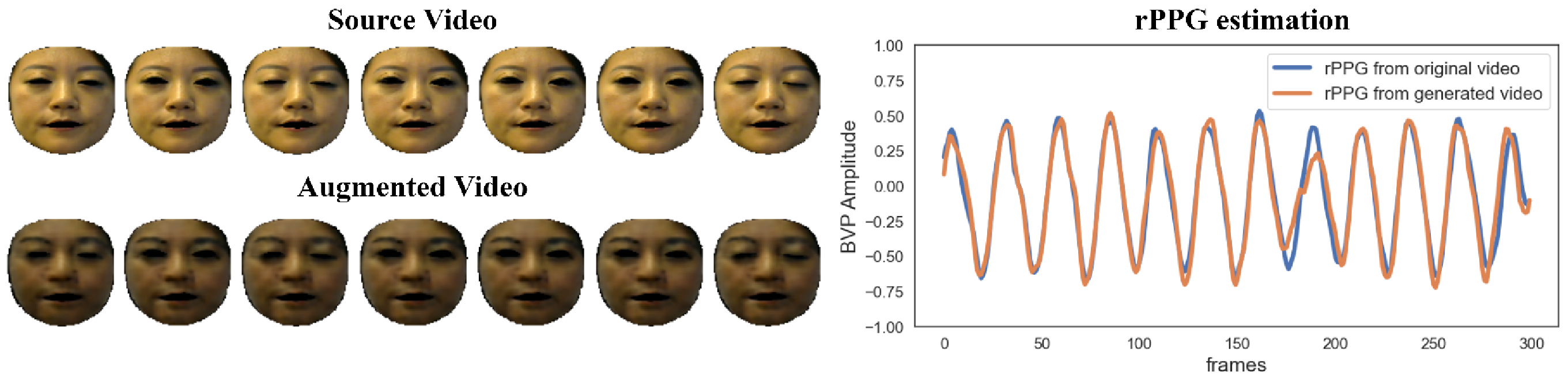}

    \caption{Visual example of dark skin tone data augmentation. PhysFlow transfers skin tone while preserving the pulsatile wave from the source to the augmented video.}
    \label{fig:qualitative}

\end{figure}

To showcase the advantages of integrating the PhysFlow approach into rPPG data-driven models for addressing biased skin tone challenges, we assess our method on skin types V and VI, which represent the most challenging skin types, as mentioned previously. To mitigate the skin tone unbalancing effect, during training PhysFlow generates an augmented range of darker skin by conditioning each sequence with a luminance value between 0.15 and 0.40, which allows the model to generate and learn from a wider diversity of darker skin tones. Fig. \ref{fig:qualitative} illustrates a qualitative demonstration of the PhysFlow augmentation for dark skin tones, while ensuring the preservation of pulsatile information extracted from the facial region.

Table \ref{table:Physflow} depicts the cross-dataset HR performance comparison among the three evaluated data-driven models for skin types V and VI in the MMPD dataset, with and without the integration of PhysFlow. The findings reveal a significant reduction in HR error, with the introduction of augmented dark skin tone videos resulting in a decrease in MAE ranging from 1 to 5 BPM, depending on the method. %Notably, skin type VI demonstrates a more substantial improvement compared to skin type V. This can be attributed to the fact that skin type VI represents a more distinct skin tone compared to the predominant skin type in the training dataset. Consequently, skin V also experiences a greater improvement but slightly less, which can be explained since skin type V is closer to the predominant skin type III in rPPG datasets. 
Skin type VI demonstrates a more substantial improvement compared to skin type V, attributed to its distinctiveness from the predominant skin type in the training dataset. While skin type V also experiences improvement, it is slightly less pronounced due to its proximity to the predominant skin type III in rPPG datasets. Additionally, there are variations in improvement among the adopted methods. For instance, the HR error for the EfficientPhys model is reduced by approximately 36 \% for skin types V and VI, whereas the DPMN+TDM model shows reductions ranging from 13\% to 22\%. This can be explained since EfficientPhys was designed as a lightweight model for rPPG deployment which due to its limited characteristics is more sensitive to skin tone bias than PhysNet or DPMN+TDM. Nevertheless, after the PhysFlow data augmentation, the HR error reduction is observed across all tested models, and their performance aligns more closely with the tendency observed for skin type III in Table \ref{table:baseline}.

\begin{table}[t]

  \renewcommand{\arraystretch}{1.25}
  \centering
  \adjustbox{width=0.75\textwidth}{
  \large
  \begin{tabular}{c|c|cccc|cccc}
    \hline
    \multicolumn{2}{c|}{Skin Tone} & \multicolumn{4}{c|}{V}& \multicolumn{4}{c}{VI}\\
    \hline
     Method & Strategy &
    MAE$\downarrow$ & RMSE$\downarrow$ & MAPE$\downarrow$ & R$\uparrow$& MAE$\downarrow$ & RMSE$\downarrow$ & MAPE$\downarrow$ & R$\uparrow$\\ 
    \hline
    \multirow{2}{*}{{\parbox{3cm}{\centering PhysNet \cite{Yu2019RemotePS} }}} 
    
    &Baseline & 10.65 & 13.73 & 15.66 & 0.16  & 13.74 & 18.19 & 16.18 & -0.04\\ 
    &PhysFlow & \textbf{7.35} & \textbf{11.71} & \textbf{11.73} & \textbf{0.47} & \textbf{9.47} & \textbf{13.72} & \textbf{11.35} & \textbf{0.40} \\ 
    \cline{2-10}
    &Difference & \color{ForestGreen} \textbf{-3.30} & \color{ForestGreen} \textbf{-2.02} & \color{ForestGreen} \textbf{-3.93} & \color{ForestGreen} \textbf{+0.31} & \color{ForestGreen} \textbf{-4.27} & \color{ForestGreen} \textbf{-4.47} & \color{ForestGreen} \textbf{-4.83} & \color{ForestGreen} \textbf{+0.43}\\
    \hline
    \multirow{2}{*}{\parbox{3cm}{\centering EfficientPhys \cite{liu2023efficientphys}}}
    & Baseline & 11.92 & 19.84 & 18.64 & 0.04 &  16.25 & 23.84 & 21.22 & 0.05\\ 
    &PhysFlow & \textbf{7.71} & \textbf{13.36} & \textbf{11.51} & \textbf{0.49}& \textbf{10.53} & \textbf{16.22} & \textbf{13.69} & \textbf{0.36}\\ 
    \cline{2-10}
    &Difference & \color{ForestGreen} \textbf{-4.21} & \color{ForestGreen} \textbf{-6.48} & \color{ForestGreen} \textbf{-7.13} & \color{ForestGreen} \textbf{+0.45} & \color{ForestGreen} \textbf{-5.72} & \color{ForestGreen} \textbf{-7.62} & \color{ForestGreen} \textbf{-7.53} & \color{ForestGreen} \textbf{+0.31}\\
    \hline
    \multirow{2}{*}{\parbox{3cm}{\centering 
    DPMN+TDM \cite{comas2024pulse}}}
    & Baseline & 7.77 & 11.90 & 11.64 & 0.46 & 11.08 & 15.22 & 13.48 & 0.25\\ 
    &PhysFlow & \textbf{6.76} & \textbf{10.33} & \textbf{10.27} & \textbf{0.63} & \textbf{8.72} & \textbf{12.32} & \textbf{10.63} & \textbf{0.47}\\ 
    \cline{2-10}
    &Difference & \color{ForestGreen} \textbf{-1.01} & \color{ForestGreen} \textbf{-1.57} & \color{ForestGreen} \textbf{-1.37} & \color{ForestGreen} \textbf{+0.17} & \color{ForestGreen} \textbf{-2.36} & \color{ForestGreen} \textbf{-2.90} & \color{ForestGreen} \textbf{-2.85} & \color{ForestGreen} \textbf{+0.22}\\
    \hline
  \end{tabular}}
   \caption{PhysFlow cross-evaluation on darkness skin types of MMPD dataset (beats per minute).}
  \label{table:Physflow}

\end{table}

\section{Conclusions and future work}
\label{sec:conclusions}

In this work, we introduce PhysFlow, a novel method for augmenting skin tone diversity in remote heart rate estimation.  PhysFlow utilizes conditional normalizing flows to disentangle skin tone information from other facial video content. Physflow is trained end-to-end, allowing simultaneous training of any supervised rPPG approach on original and augmented data during data generation. To train PhysFlow we use a novel bi-dimensional skin tone representation using CIELAB color space, offering adaptability to unbalanced rPPG datasets and skin tone variations. Our cross-dataset experiments on the MMPD dataset using three different data-driven models demonstrate the capability of PhysFlow for skin tone diversity augmenting in any supervised rPPG, showing how our approach significantly reduces heart rate estimation error, particularly in underrepresented skin tone categories, favoring equitable performance across different skin tones. Future work will focus on extending the data augmentation process to include the generation of new appearance data and modification of pulsatile information through a multi-modal conditional normalizing flow approach. 
\section*{Acknowledgments}
This work is partly supported by the eSCANFace project (PID2020-114083GB-I00) funded by the Spanish Ministry of Science and Innovation.

%%%%%%%%% REFERENCES
\bibliography{References}

\begin{thebibliography}{65}
\providecommand{\natexlab}[1]{#1}
\providecommand{\url}[1]{\texttt{#1}}
\expandafter\ifx\csname urlstyle\endcsname\relax
  \providecommand{\doi}[1]{doi: #1}\else
  \providecommand{\doi}{doi: \begingroup \urlstyle{rm}\Url}\fi

\bibitem[Abdal et~al.(2021)Abdal, Zhu, Mitra, and Wonka]{abdal2021styleflow}
Rameen Abdal, Peihao Zhu, Niloy~J Mitra, and Peter Wonka.
\newblock Styleflow: Attribute-conditioned exploration of stylegan-generated images using conditional continuous normalizing flows.
\newblock \emph{ACM Transactions on Graphics (ToG)}, 40\penalty0 (3):\penalty0 1--21, 2021.

\bibitem[Addison et~al.(2018)Addison, Jacquel, Foo, and Borg]{addison2018video}
Paul~S Addison, Dominique Jacquel, David~MH Foo, and Ulf~R Borg.
\newblock Video-based heart rate monitoring across a range of skin pigmentations during an acute hypoxic challenge.
\newblock \emph{Journal of clinical monitoring and computing}, 32:\penalty0 871--880, 2018.

\bibitem[Ba et~al.(2022)Ba, Wang, Karinca, Bozkurt, and Kadambi]{ba2022style}
Yunhao Ba, Zhen Wang, Kerim~Doruk Karinca, Oyku~Deniz Bozkurt, and Achuta Kadambi.
\newblock Style transfer with bio-realistic appearance manipulation for skin-tone inclusive rppg.
\newblock In \emph{2022 IEEE International Conference on Computational Photography (ICCP)}, pages 1--12. IEEE, 2022.

\bibitem[Benezeth et~al.(2018)Benezeth, Li, Macwan, Nakamura, Gomez, and Yang]{benezeth2018remote}
Yannick Benezeth, Peixi Li, Richard Macwan, Keisuke Nakamura, Randy Gomez, and Fan Yang.
\newblock Remote heart rate variability for emotional state monitoring.
\newblock In \emph{EMBS Int. Conf. Biomed. Health Inform. BHI}, pages 153--156. IEEE, 2018.

\bibitem[Chen et~al.(2018)Chen, Rubanova, Bettencourt, and Duvenaud]{chen2018neural}
Ricky~TQ Chen, Yulia Rubanova, Jesse Bettencourt, and David~K Duvenaud.
\newblock Neural ordinary differential equations.
\newblock \emph{Advances in neural information processing systems}, 31, 2018.

\bibitem[Chen and McDuff(2018)]{chen2018deepphys}
Weixuan Chen and Daniel McDuff.
\newblock Deepphys: Video-based physiological measurement using convolutional attention networks.
\newblock In \emph{ECCV}, pages 349--365, 2018.

\bibitem[Comas et~al.(2022)Comas, Ruiz, and Sukno]{comas2022efficient}
Joaquim Comas, Adria Ruiz, and Federico Sukno.
\newblock Efficient remote photoplethysmography with temporal derivative modules and time-shift invariant loss.
\newblock In \emph{CVPR}, pages 2182--2191, 2022.

\bibitem[Comas et~al.(2024{\natexlab{a}})Comas, Ruiz, and Sukno]{comas2024deep}
Joaquim Comas, Adria Ruiz, and Federico Sukno.
\newblock Deep adaptative spectral zoom for improved remote heart rate estimation.
\newblock \emph{arXiv preprint arXiv:2403.06902}, 2024{\natexlab{a}}.

\bibitem[Comas et~al.(2024{\natexlab{b}})Comas, Ruiz, and Sukno]{comas2024pulse}
Joaquim Comas, Adria Ruiz, and Federico Sukno.
\newblock Deep pulse-signal magnification for remote heart rate estimation in compressed videos.
\newblock \emph{arXiv preprint arXiv:2405.02652}, 2024{\natexlab{b}}.

\bibitem[Dall’Olio et~al.(2020)Dall’Olio, Curti, Remondini, Safi~Harb, Asselbergs, Castellani, and Uh]{dall2020prediction}
Lorenzo Dall’Olio, Nico Curti, Daniel Remondini, Yosef Safi~Harb, Folkert~W Asselbergs, Gastone Castellani, and Hae-Won Uh.
\newblock Prediction of vascular aging based on smartphone acquired ppg signals.
\newblock \emph{Scientific reports}, 10:\penalty0 1--10, 2020.

\bibitem[Dasari et~al.(2021)Dasari, Prakash, Jeni, and Tucker]{dasari2021evaluation}
Ananyananda Dasari, Sakthi Kumar~Arul Prakash, L{\'a}szl{\'o}~A Jeni, and Conrad~S Tucker.
\newblock Evaluation of biases in remote photoplethysmography methods.
\newblock \emph{NPJ digital medicine}, 4\penalty0 (1):\penalty0 91, 2021.

\bibitem[De~Haan and Jeanne(2013)]{de2013robust}
Gerard De~Haan and Vincent Jeanne.
\newblock Robust pulse rate from chrominance-based rppg.
\newblock \emph{IEEE Trans. Biomed. Eng.}, 60\penalty0 (10):\penalty0 2878--2886, 2013.

\bibitem[Dinh et~al.(2015)Dinh, Krueger, and Bengio]{dinh2014nice}
Laurent Dinh, David Krueger, and Yoshua Bengio.
\newblock Nice: Non-linear independent components estimation.
\newblock \emph{ICLR}, 2015.

\bibitem[Dinh et~al.(2017)Dinh, Sohl-Dickstein, and Bengio]{dinh2017density}
Laurent Dinh, Jascha Sohl-Dickstein, and Samy Bengio.
\newblock Density estimation using real {NVP}.
\newblock In \emph{ICLR}, 2017.

\bibitem[Ernst et~al.(2021)Ernst, Scherpf, Malberg, and Schmidt]{ernst2021optimal}
Hannes Ernst, Matthieu Scherpf, Hagen Malberg, and Martin Schmidt.
\newblock Optimal color channel combination across skin tones for remote heart rate measurement in camera-based photoplethysmography.
\newblock \emph{Biomedical Signal Processing and Control}, 68:\penalty0 102644, 2021.

\bibitem[Fallow et~al.(2013)Fallow, Tarumi, and Tanaka]{fallow2013influence}
Bennett~A Fallow, Takashi Tarumi, and Hirofumi Tanaka.
\newblock Influence of skin type and wavelength on light wave reflectance.
\newblock \emph{Journal of clinical monitoring and computing}, 27:\penalty0 313--317, 2013.

\bibitem[Fitzpatrick(1988)]{fitzpatrick1988validity}
Thomas~B Fitzpatrick.
\newblock The validity and practicality of sun-reactive skin types i through vi.
\newblock \emph{Archives of dermatology}, 124\penalty0 (6):\penalty0 869--871, 1988.

\bibitem[Grathwohl et~al.(2019)Grathwohl, Chen, Bettencourt, and Duvenaud]{grathwohl2018scalable}
Will Grathwohl, Ricky T.~Q. Chen, Jesse Bettencourt, and David Duvenaud.
\newblock Scalable reversible generative models with free-form continuous dynamics.
\newblock In \emph{International Conference on Learning Representations}, 2019.
\newblock URL \url{https://openreview.net/forum?id=rJxgknCcK7}.

\bibitem[Gupta et~al.(2023)Gupta, Kumar, Birla, and Gupta]{gupta2023radiant}
Anup~Kumar Gupta, Rupesh Kumar, Lokendra Birla, and Puneet Gupta.
\newblock Radiant: Better rppg estimation using signal embeddings and transformer.
\newblock In \emph{Proceedings of the IEEE/CVF Winter Conference on Applications of Computer Vision}, pages 4976--4986, 2023.

\bibitem[Hsieh et~al.(2022)Hsieh, Chung, and Hsu]{hsieh2022augmentation}
Cheng-Ju Hsieh, Wei-Hao Chung, and Chiou-Ting Hsu.
\newblock Augmentation of rppg benchmark datasets: Learning to remove and embed rppg signals via double cycle consistent learning from unpaired facial videos.
\newblock In \emph{European Conference on Computer Vision}, pages 372--387. Springer, 2022.

\bibitem[Huang et~al.(2021)Huang, Chen, Lin, Juang, Xing, Wang, and Wang]{huang2021neonatal}
Bin Huang, Weihai Chen, Chun-Liang Lin, Chia-Feng Juang, Yuanping Xing, Yanting Wang, and Jianhua Wang.
\newblock A neonatal dataset and benchmark for non-contact neonatal heart rate monitoring based on spatio-temporal neural networks.
\newblock \emph{Engineering Applications of Artificial Intelligence}, 106:\penalty0 104447, 2021.

\bibitem[Johnson et~al.(2016)Johnson, Alahi, and Fei-Fei]{johnson2016perceptual}
Justin Johnson, Alexandre Alahi, and Li~Fei-Fei.
\newblock Perceptual losses for real-time style transfer and super-resolution.
\newblock In \emph{Computer Vision--ECCV 2016: 14th European Conference, Amsterdam, The Netherlands, October 11-14, 2016, Proceedings, Part II 14}, pages 694--711. Springer, 2016.

\bibitem[Kartynnik et~al.(2019)Kartynnik, Ablavatski, Grishchenko, and Grundmann]{kartynnik2019real}
Yury Kartynnik, Artsiom Ablavatski, Ivan Grishchenko, and Matthias Grundmann.
\newblock Real-time facial surface geometry from monocular video on mobile gpus.
\newblock \emph{arXiv preprint arXiv:1907.06724}, 2019.

\bibitem[Lee et~al.(2020)Lee, Chen, and Lee]{lee2020meta}
Eugene Lee, Evan Chen, and Chen-Yi Lee.
\newblock Meta-rppg: Remote heart rate estimation using a transductive meta-learner.
\newblock In \emph{ECCV}, pages 392--409. Springer, 2020.

\bibitem[Lewandowska et~al.(2011)Lewandowska, Rumi{\'n}ski, Kocejko, and Nowak]{lewandowska2011measuring}
Magdalena Lewandowska, Jacek Rumi{\'n}ski, Tomasz Kocejko, and J{e}drzej Nowak.
\newblock Measuring pulse rate with a webcam—a non-contact method for evaluating cardiac activity.
\newblock In \emph{FedCSIS}, pages 405--410. IEEE, 2011.

\bibitem[Li et~al.(2014)Li, Chen, Zhao, and Pietikainen]{li2014remote}
Xiaobai Li, Jie Chen, Guoying Zhao, and Matti Pietikainen.
\newblock Remote heart rate measurement from face videos under realistic situations.
\newblock In \emph{CVPR}, pages 4264--4271, 2014.

\bibitem[Li et~al.(2022)Li, Lien, and Wang]{li2022style}
Yuan-kui Li, Yun-Hsuan Lien, and Yu-Shuen Wang.
\newblock Style-structure disentangled features and normalizing flows for diverse icon colorization.
\newblock In \emph{Proceedings of the IEEE/CVF Conference on Computer Vision and Pattern Recognition}, pages 11244--11253, 2022.

\bibitem[Lippe et~al.(2022)Lippe, Magliacane, L{\"o}we, Asano, Cohen, and Gavves]{lippe2022citris}
Phillip Lippe, Sara Magliacane, Sindy L{\"o}we, Yuki~M Asano, Taco Cohen, and Stratis Gavves.
\newblock Citris: Causal identifiability from temporal intervened sequences.
\newblock In \emph{International Conference on Machine Learning}, pages 13557--13603. PMLR, 2022.

\bibitem[Liu et~al.(2021)Liu, Jiang, Fromm, Xu, Patel, and McDuff]{liu2021metaphys}
Xin Liu, Ziheng Jiang, Josh Fromm, Xuhai Xu, Shwetak Patel, and Daniel McDuff.
\newblock Metaphys: few-shot adaptation for non-contact physiological measurement.
\newblock In \emph{CHIL}, pages 154--163, 2021.

\bibitem[Liu et~al.(2023)Liu, Hill, Jiang, Patel, and McDuff]{liu2023efficientphys}
Xin Liu, Brian Hill, Ziheng Jiang, Shwetak Patel, and Daniel McDuff.
\newblock Efficientphys: Enabling simple, fast and accurate camera-based cardiac measurement.
\newblock In \emph{IEEE Winter Conf. Appl. Comput. Vis.}, pages 5008--5017, 2023.

\bibitem[Lu et~al.(2021)Lu, Han, and Zhou]{lu2021dual}
Hao Lu, Hu~Han, and S~Kevin Zhou.
\newblock Dual-gan: Joint bvp and noise modeling for remote physiological measurement.
\newblock In \emph{CVPR}, pages 12404--12413, 2021.

\bibitem[Lugaresi et~al.(2019)Lugaresi, Tang, Nash, McClanahan, Uboweja, Hays, Zhang, Chang, Yong, Lee, et~al.]{lugaresi2019mediapipe}
Camillo Lugaresi, Jiuqiang Tang, Hadon Nash, Chris McClanahan, Esha Uboweja, Michael Hays, Fan Zhang, Chuo-Ling Chang, Ming~Guang Yong, Juhyun Lee, et~al.
\newblock Mediapipe: A framework for building perception pipelines.
\newblock \emph{arXiv preprint arXiv:1906.08172}, 2019.

\bibitem[McDuff et~al.(2022)McDuff, Wander, Liu, Hill, Hernandez, Lester, and Baltrusaitis]{mcduff2022scamps}
Daniel McDuff, Miah Wander, Xin Liu, Brian Hill, Javier Hernandez, Jonathan Lester, and Tadas Baltrusaitis.
\newblock Scamps: Synthetics for camera measurement of physiological signals.
\newblock \emph{Advances in Neural Information Processing Systems}, 35:\penalty0 3744--3757, 2022.

\bibitem[Niu et~al.(2018{\natexlab{a}})Niu, Han, Shan, and Chen]{niu2018synrhythm}
Xuesong Niu, Hu~Han, Shiguang Shan, and Xilin Chen.
\newblock Synrhythm: Learning a deep heart rate estimator from general to specific.
\newblock In \emph{ICPR}, pages 3580--3585. IEEE, 2018{\natexlab{a}}.

\bibitem[Niu et~al.(2018{\natexlab{b}})Niu, Han, Shan, and Chen]{niu2018vipl}
Xuesong Niu, Hu~Han, Shiguang Shan, and Xilin Chen.
\newblock Vipl-hr: A multi-modal database for pulse estimation from less-constrained face video.
\newblock In \emph{ACCV}, pages 562--576. Springer, 2018{\natexlab{b}}.

\bibitem[Niu et~al.(2019)Niu, Shan, Han, and Chen]{niu2019rhythmnet}
Xuesong Niu, Shiguang Shan, Hu~Han, and Xilin Chen.
\newblock Rhythmnet: End-to-end heart rate estimation from face via spatial-temporal representation.
\newblock \emph{IEEE TIP}, 29:\penalty0 2409--2423, 2019.

\bibitem[Nowara et~al.(2020{\natexlab{a}})Nowara, Marks, Mansour, and Veeraraghavan]{nowara2020near}
Ewa~M Nowara, Tim~K Marks, Hassan Mansour, and Ashok Veeraraghavan.
\newblock Near-infrared imaging photoplethysmography during driving.
\newblock \emph{IEEE transactions on intelligent transportation systems}, 23\penalty0 (4):\penalty0 3589--3600, 2020{\natexlab{a}}.

\bibitem[Nowara et~al.(2020{\natexlab{b}})Nowara, McDuff, and Veeraraghavan]{nowara2020meta}
Ewa~M Nowara, Daniel McDuff, and Ashok Veeraraghavan.
\newblock A meta-analysis of the impact of skin tone and gender on non-contact photoplethysmography measurements.
\newblock In \emph{Proceedings of the IEEE/CVF Conference on Computer Vision and Pattern Recognition Workshops}, pages 284--285, 2020{\natexlab{b}}.

\bibitem[Nowara et~al.(2021)Nowara, McDuff, and Veeraraghavan]{nowara2021benefit}
Ewa~M Nowara, Daniel McDuff, and Ashok Veeraraghavan.
\newblock The benefit of distraction: Denoising camera-based physiological measurements using inverse attention.
\newblock In \emph{ICCV}, pages 4955--4964, 2021.

\bibitem[Paruchuri et~al.(2024)Paruchuri, Liu, Pan, Patel, McDuff, and Sengupta]{paruchuri2024motion}
Akshay Paruchuri, Xin Liu, Yulu Pan, Shwetak Patel, Daniel McDuff, and Soumyadip Sengupta.
\newblock Motion matters: Neural motion transfer for better camera physiological measurement.
\newblock In \emph{Proceedings of the IEEE/CVF Winter Conference on Applications of Computer Vision}, pages 5933--5942, 2024.

\bibitem[Paszke et~al.(2019)Paszke, Gross, Massa, Lerer, Bradbury, Chanan, Killeen, Lin, Gimelshein, Antiga, et~al.]{paszke2019pytorch}
Adam Paszke, Sam Gross, Francisco Massa, Adam Lerer, James Bradbury, Gregory Chanan, Trevor Killeen, Zeming Lin, Natalia Gimelshein, Luca Antiga, et~al.
\newblock Pytorch: An imperative style, high-performance deep learning library.
\newblock \emph{Adv Neural Inf Process Syst}, 32, 2019.

\bibitem[Perepelkina et~al.(2020)Perepelkina, Artemyev, Churikova, and Grinenko]{perepelkina2020hearttrack}
Olga Perepelkina, Mikhail Artemyev, Marina Churikova, and Mikhail Grinenko.
\newblock Hearttrack: Convolutional neural network for remote video-based heart rate monitoring.
\newblock In \emph{CVPRW}, pages 288--289, 2020.

\bibitem[Poh et~al.(2010{\natexlab{a}})Poh, McDuff, and Picard]{poh2010advancements}
Ming-Zher Poh, Daniel~J McDuff, and Rosalind~W Picard.
\newblock Advancements in noncontact, multiparameter physiological measurements using a webcam.
\newblock \emph{IEEE Trans. Biomed. Eng.}, 58\penalty0 (1):\penalty0 7--11, 2010{\natexlab{a}}.

\bibitem[Poh et~al.(2010{\natexlab{b}})Poh, McDuff, and Picard]{poh2010non}
Ming-Zher Poh, Daniel~J McDuff, and Rosalind~W Picard.
\newblock Non-contact, automated cardiac pulse measurements using video imaging and blind source separation.
\newblock \emph{Optics express}, 18\penalty0 (10):\penalty0 10762--10774, 2010{\natexlab{b}}.

\bibitem[Pontryagin(2018)]{pontryagin2018mathematical}
Lev~Semenovich Pontryagin.
\newblock \emph{Mathematical theory of optimal processes}.
\newblock Routledge, 2018.

\bibitem[Rezende and Mohamed(2015)]{rezende2015variational}
Danilo Rezende and Shakir Mohamed.
\newblock Variational inference with normalizing flows.
\newblock In \emph{International conference on machine learning}, pages 1530--1538. PMLR, 2015.

\bibitem[Song et~al.(2021)Song, Chen, Cheng, Li, Liu, and Chen]{song2021pulsegan}
Rencheng Song, Huan Chen, Juan Cheng, Chang Li, Yu~Liu, and Xun Chen.
\newblock Pulsegan: Learning to generate realistic pulse waveforms in remote photoplethysmography.
\newblock \emph{IEEE J.Biomed.Health Inform.}, 25\penalty0 (5):\penalty0 1373--1384, 2021.

\bibitem[{\v{S}}petl{\'\i}k et~al.(2018){\v{S}}petl{\'\i}k, Franc, and Matas]{vspetlik2018visual}
Radim {\v{S}}petl{\'\i}k, Vojtech Franc, and Jir{\'\i} Matas.
\newblock Visual heart rate estimation with convolutional neural network.
\newblock In \emph{BMVC}, 2018.

\bibitem[Stricker et~al.(2014)Stricker, M{\"u}ller, and Gross]{stricker2014non}
Ronny Stricker, Steffen M{\"u}ller, and Horst-Michael Gross.
\newblock Non-contact video-based pulse rate measurement on a mobile service robot.
\newblock In \emph{RO-MAN}, pages 1056--1062. IEEE, 2014.

\bibitem[Tabak and Vanden-Eijnden(2010)]{tabak2010density}
Esteban~G Tabak and Eric Vanden-Eijnden.
\newblock Density estimation by dual ascent of the log-likelihood.
\newblock \emph{Communications in Mathematical Sciences}, 8\penalty0 (1):\penalty0 217--233, 2010.

\bibitem[Takano and Ohta(2007)]{takano2007heart}
Chihiro Takano and Yuji Ohta.
\newblock Heart rate measurement based on a time-lapse image.
\newblock \emph{Medical engineering \& physics}, 29\penalty0 (8):\penalty0 853--857, 2007.

\bibitem[Tang et~al.(2023)Tang, Chen, Wang, Shi, Patel, McDuff, and Liu]{tang2023mmpd}
Jiankai Tang, Kequan Chen, Yuntao Wang, Yuanchun Shi, Shwetak~N. Patel, Daniel~J. McDuff, and Xin Liu.
\newblock Mmpd: Multi-domain mobile video physiology dataset.
\newblock \emph{2023 45th Annual International Conference of the IEEE Engineering in Medicine \& Biology Society (EMBC)}, pages 1--5, 2023.
\newblock URL \url{https://api.semanticscholar.org/CorpusID:256662570}.

\bibitem[Thong et~al.(2023)Thong, Joniak, and Xiang]{thong2023beyond}
William Thong, Przemyslaw Joniak, and Alice Xiang.
\newblock Beyond skin tone: A multidimensional measure of apparent skin color.
\newblock In \emph{Proceedings of the IEEE/CVF International Conference on Computer Vision}, pages 4903--4913, 2023.

\bibitem[Tsou et~al.(2020)Tsou, Lee, and Hsu]{tsou2020multi}
Yun-Yun Tsou, Yi-An Lee, and Chiou-Ting Hsu.
\newblock Multi-task learning for simultaneous video generation and remote photoplethysmography estimation.
\newblock In \emph{ACCV}, 2020.

\bibitem[Tulyakov et~al.(2016)Tulyakov, Alameda-Pineda, Ricci, Yin, Cohn, and Sebe]{tulyakov2016self}
Sergey Tulyakov, Xavier Alameda-Pineda, Elisa Ricci, Lijun Yin, Jeffrey~F Cohn, and Nicu Sebe.
\newblock Self-adaptive matrix completion for heart rate estimation from face videos under realistic conditions.
\newblock In \emph{CVPR}, pages 2396--2404, 2016.

\bibitem[Verkruysse et~al.(2008)Verkruysse, Svaasand, and Nelson]{verkruysse2008remote}
Wim Verkruysse, Lars~O Svaasand, and J~Stuart Nelson.
\newblock Remote plethysmographic imaging using ambient light.
\newblock \emph{Optics express}, 16\penalty0 (26):\penalty0 21434--21445, 2008.

\bibitem[Vilesov et~al.(2022)Vilesov, Chari, Armouti, Harish, Kulkarni, Deoghare, Jalilian, and Kadambi]{vilesov2022blending}
Alexander Vilesov, Pradyumna Chari, Adnan Armouti, Anirudh~Bindiganavale Harish, Kimaya Kulkarni, Ananya Deoghare, Laleh Jalilian, and Achuta Kadambi.
\newblock Blending camera and 77 ghz radar sensing for equitable, robust plethysmography.
\newblock \emph{ACM Trans. Graph.}, 41\penalty0 (4):\penalty0 36--1, 2022.

\bibitem[Wang and Deng(2018)]{wang2018deep}
Mei Wang and Weihong Deng.
\newblock Deep visual domain adaptation: A survey.
\newblock \emph{Neurocomputing}, 312:\penalty0 135--153, 2018.

\bibitem[Wang et~al.(2015)Wang, Stuijk, and De~Haan]{wang2015novel}
Wenjin Wang, Sander Stuijk, and Gerard De~Haan.
\newblock A novel algorithm for remote photoplethysmography: Spatial subspace rotation.
\newblock \emph{IEEE Trans. Biomed. Eng.}, 63\penalty0 (9):\penalty0 1974--1984, 2015.

\bibitem[Wang et~al.(2016)Wang, den Brinker, Stuijk, and De~Haan]{wang2016algorithmic}
Wenjin Wang, Albertus~C den Brinker, Sander Stuijk, and Gerard De~Haan.
\newblock Algorithmic principles of remote ppg.
\newblock \emph{IEEE Trans. Biomed. Eng.}, 64\penalty0 (7):\penalty0 1479--1491, 2016.

\bibitem[Wang et~al.(2022)Wang, Ba, Chari, Bozkurt, Brown, Patwa, Vaddi, Jalilian, and Kadambi]{wang2022synthetic}
Zhen Wang, Yunhao Ba, Pradyumna Chari, Oyku~Deniz Bozkurt, Gianna Brown, Parth Patwa, Niranjan Vaddi, Laleh Jalilian, and Achuta Kadambi.
\newblock Synthetic generation of face videos with plethysmograph physiology.
\newblock In \emph{CVPR}, pages 20587--20596, 2022.

\bibitem[Yu et~al.(2019{\natexlab{a}})Yu, Li, and Zhao]{Yu2019RemotePS}
Z.~Yu, Xiao-Bai Li, and G.~Zhao.
\newblock Remote photoplethysmograph signal measurement from facial videos using spatio-temporal networks.
\newblock In \emph{BMVC}, 2019{\natexlab{a}}.

\bibitem[Yu et~al.(2019{\natexlab{b}})Yu, Peng, Li, Hong, and Zhao]{yu2019remote}
Zitong Yu, Wei Peng, Xiaobai Li, Xiaopeng Hong, and Guoying Zhao.
\newblock Remote heart rate measurement from highly compressed facial videos: an end-to-end deep learning solution with video enhancement.
\newblock In \emph{ICCV}, pages 151--160, 2019{\natexlab{b}}.

\bibitem[Yu et~al.(2023)Yu, Shen, Shi, Zhao, Cui, Zhang, Torr, and Zhao]{yu2023physformer++}
Zitong Yu, Yuming Shen, Jingang Shi, Hengshuang Zhao, Yawen Cui, Jiehua Zhang, Philip Torr, and Guoying Zhao.
\newblock Physformer++: Facial video-based physiological measurement with slowfast temporal difference transformer.
\newblock \emph{IJCV}, 131\penalty0 (6):\penalty0 1307--1330, 2023.

\bibitem[Yucer et~al.(2020)Yucer, Ak{\c{c}}ay, Al-Moubayed, and Breckon]{yucer2020exploring}
Seyma Yucer, Samet Ak{\c{c}}ay, Noura Al-Moubayed, and Toby~P Breckon.
\newblock Exploring racial bias within face recognition via per-subject adversarially-enabled data augmentation.
\newblock In \emph{Proceedings of the IEEE/CVF Conference on Computer Vision and Pattern Recognition Workshops}, pages 18--19, 2020.

\end{thebibliography}
\end{document}